\documentclass[letterpaper, 10 pt, conference]{ieeeconf}  
\IEEEoverridecommandlockouts                               
\usepackage{graphicx}                       
\usepackage{graphics}                       
\usepackage{epsfig}                         
\usepackage[tight,footnotesize]{subfigure}  
\usepackage{amssymb,amsmath}
\usepackage{mdwmath}
\usepackage{commath}                        
\usepackage{eqparbox}
\usepackage{amsmath,amsfonts,amssymb}

\newcommand{\ie}{{\em i.e.\ }}
\newcommand{\Ie}{{\em I.e.\ }}

\newcommand{\et}{{\em et al.\ }}

\newcommand{\gap}{\mbox{ }}

\newcommand{\BC}{\mbox{$\mathcal B$}}

\newcommand{\EC}{\mbox{$\mathcal E$}}

\newcommand{\GC}{\mbox{$\mathcal G$}}

\newcommand{\NC}{\mbox{$\mathcal N$}}

\newcommand{\SC}{\mbox{$\mathcal S$}}

\newcommand{\XC}{\mbox{$\mathcal X$}}

\newcommand{\beq}{\begin{equation}}
\newcommand{\eeq}{\end{equation}}
\newcommand{\bear}{\begin{eqnarray}}
\newcommand{\bears}{\begin{eqnarray*}}
\newcommand{\eear}{\end{eqnarray}}
\newcommand{\eears}{\end{eqnarray*}}
\newcommand{\bdm}{\begin{displaymath}}
\newcommand{\edm}{\end{displaymath}}
\newcommand{\lba}{\left[\begin{array}}
\newcommand{\ear}{\end{array}\right]}

\usepackage{stfloats}                       
\usepackage{hyperref}
\usepackage{cite}                           
\usepackage[T1]{fontenc} 
\usepackage{tabularx}
\usepackage[table]{xcolor}
\usepackage{longtable}
\usepackage{booktabs}       					
\usepackage{xcolor,colortbl}               		
\newcommand{\red}{\textcolor[rgb]{ .773,  0,  .043}}
\newcommand\given[1][]{\:#1\vert\:}
\title{\LARGE \bf Recovering from External Disturbances in Online Manipulation through State-Dependent Revertive Recovery Policies.}
\author{Hongmin Wu, Shuangqi Luo, Hongbin Lin, Shuangda Duan, Yisheng Guan, and Juan Rojas. \\
\thanks{Hongmin Wu and Shuangqi Luo contributed to the present work in equal parts and share first authorship. All authors are with the School of Electromechanical Engineering in Guangdong University of Technology in Guangzhou, China.}%
}
\begin{document}
\maketitle
\thispagestyle{empty}
\pagestyle{empty}
\bstctlcite{IEEEexample:BSTcontrol}
\begin{abstract}
Robots are increasingly entering uncertain and unstructured environments. Within these, robots are bound to face unexpected external disturbances like accidental human or tool collisions. Robots must develop the capacity to respond to unexpected events. That is not only identifying the sudden anomaly, but also deciding how to handle it. 
In this work, we contribute a recovery policy that allows a robot to recovery from various anomalous scenarios across different tasks and conditions in a consistent and robust fashion. 
The system organizes tasks as a sequence of nodes composed of internal modules such as motion generation and introspection. When an introspection module flags an anomaly, the recovery strategy is triggered and reverts the task execution by selecting a target node as a function of a state dependency chart. The new skill allows the robot to overcome the effects of the external disturbance and conclude the task. 
Our system recovers from accidental human and tool collisions in a number of tasks. Of particular importance is the fact that we test the robustness of the recovery system by triggering anomalies at each node in the task graph showing robust recovery everywhere in the task. We also trigger multiple and repeated anomalies at each of the nodes of the task showing that the recovery system can consistently recover anywhere in the presence of strong and pervasive anomalous conditions. Robust recovery systems will be key enablers for long-term autonomy in robot systems.
Supplemental information including videos, code, and result analysis can be found at \cite{2017Humanoids-Rojas-supplementalURL}.
\end{abstract}
\section{INTRODUCTION}\label{sec:Intro}
\begin{figure}[t]    
	\centering		
		\includegraphics[scale=1]{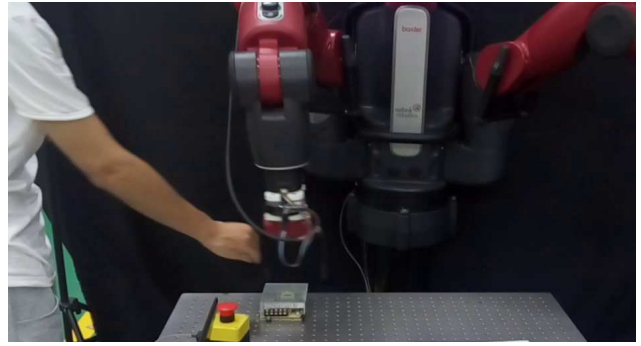}
		\caption{Illustration of an external perturbation during a pick-and-place task. A human collides with the robot arm just before the pick action leading the robot to an anomalous situation. The robot introspection system identifies the anomaly and the recovery strategy allows the robot to complete the task.}
        \label{fig:exp_setup}        
\end{figure}

%
%
Human decision making implies awareness. Adult humans are aware of their mistakes and learn to avoid making the same mistake twice. Humans also evaluate whether they have enough information before making a choice and if appropriate to proceed. Their decision's confidence is correlated with outcome success \cite{2014CNMC-Yeung-SharedMechsConfJudg_ErrDet_HumanDecisionMaking}. In robotics, online decision making and robot introspection have begun to receive more attention recently \cite{2011IROS-Rodriguez-AbortRetry,2013IROS-DiLello-BayesianContFaultDetection,2014ICRA-kroemer-LearnPredictPhasesManipHiddenStates, 2014ICRA-Rojas-EarlyFC, 2015RSS-Kappler-DateDrivenOnlineDecisionMakingManipu, 2016ICRA-Park-MultiModalMonitoringAnomalyDet_RobotManip, 2017iros-rojas-onlinewrenchintrospection}. The vision is to endow robots with the ability to understand their actions and make timely decisions to achieve their goals and have long-term autonomy. Particularly, in unstructured environments (where robots are expected to participate increasingly), external perturbations are hard to model in low-level control systems and often lead to failure. Robots must then discern nominal from anomalous conditions and trigger responses to avert failure and recover gracefully. Fig. \ref{fig:exp_setup}, illustrates an accidental collision between a human and a robot at the moment in which a robot picks an object. Normally, this situation would lead to failure, but our system enables the robot to recover and continue execution. 

This paper spans the areas of robot introspection, decision making, and anomaly recovery in robot manipulation tasks. In the literature, many works have not attempted an integral approach. Some only model success \textit{vs.} failure behaviors \cite{2014ICRA-Rojas-EarlyFC,2011IROS-Rodriguez-AbortRetry}; others do introspection or monitoring but not recovery: \cite{2013IROS-Nakamura-ErrorRecTaskStrat,2013IROS-DiLello-BayesianContFaultDetection,2016ICRA-Park-MultiModalMonitoringAnomalyDet_RobotManip,2017iros-rojas-onlinewrenchintrospection}. Kappler \et in \cite{2015RSS-Kappler-DateDrivenOnlineDecisionMakingManipu} present an integrated system for robot introspection with online decision making and some anomaly recovery. Their system was shown to be robust against a human pulling at the robot arm. However, they perform anomaly recovery at only one stage of a single task and do not provide quantitative or qualitative analysis of their method (Sec. \ref{sec:lit_review}). 

Our work studies the feasibility, speed, and robustness of a recovery strategy that enables recovery from disturbances like accidental human or tool collisions. We also study if the robot can recover, not only at a single incident in the task, but at different points in the task. There seem to be no studies that examine and test anomalous recovery robustly at multiple points in the task. Finally, we study the robot's ability not only to recover at multiple points in a task, but also in situations where a disturbance happens repeatedly at the same point in a task. That is, once the robot recovers, if forced again into an anomalous situation, can the robot recover repeatedly? \Ie can there be a smooth and continuous ability to re-set and re-start a task? Our contribution is the implementation of a fast and robust recovery strategy that allows recovery from one or multiple anomalous situations throughout the task. 

\begin{figure}     
	\centering		
		\includegraphics[scale=0.9]{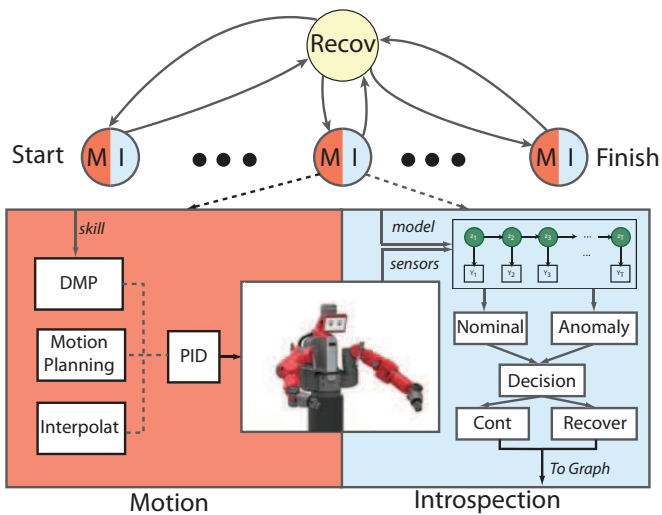}
		\caption{Our robot introspection framework for decision making in anomaly recovery organizes specific tasks as a directed graph. Each node in the graph enacts a learned: (i) manipulation skill and (ii) introspection model based on the nominal sensory-signals for such a skill. If an anomaly is present, the system triggers a generic recovery policy that reverts the system according to a node-dependency criteria.}
        \label{fig:framework}        
\end{figure}

Tasks are modeled as a directed graphical model, where nodes within the graph play a dual role: it establishes both a robot skill execution and a trained model for robot introspection (based on nonparametric Bayesian Markov switching processes, see Sec. \ref{sec:robot_introspection}). Node connections define successor nodes. During the execution of a node, the robot introspection system identifies the current skill and any anomalies that may occur therein (Sec. \ref{subsec:identification}). If an anomaly is identified, a generic recovery strategy is enacted. We opted for a generic strategy that works despite commonly large anomaly spaces that are hard to identify. Our recovery strategy reverts the task execution to a target node that is selected according to a node-dependency criteria (Sec. \ref{sec:recovery}). The recovery system is tightly connected to the lower-level control system, issuing commands to recover or allowing the system to continue. The framework is presented in Fig. \ref{fig:framework}.

We measure the robustness and flexibility of the recovery strategy in a pick-and-place experiment as well as a open-and-close drawer experiment, both of which where subjected to external disturbances like human collisions and tool collisions. Anomalies were induced under different conditions; namely: one anomaly caused during each of the nodes of a task, and a repeated number of anomalies caused during each node of the task.
F-score, micro- and macro- precision-recall statistics are computed for all experiments and conditions, where F-scores indicates the robot's recovery rate. An average recovery rate of 88.18\% was achieved for all experiments and conditions indicating the utility of the a recovery strategy on top of an introspection system to increase robustness and extend autonomy for robot systems in unstructured environments.
Robots are still prone to frequent failure in unstructured environments. Unless they can gracefully recover in a predictable fashion, it will remain challenging for humans to accept robots as reliable partners.
Supplemental information, code, data and videos can be found at \cite{2017Humanoids-Rojas-supplementalURL}.
\section{Literature Review} \label{sec:lit_review}
Our work integrates robot introspection, decision making, and anomaly recovery in robot manipulation tasks. Past works mostly focus on single issues with few works attempting integrated solutions. 
In \cite{2011IROS-Rodriguez-AbortRetry}, Rodriguez \et, designed an ``Abort and Retry'' solution to the problem of bin-picking using a simple hand. This is an early work that models \textit{success vs. failure classification} and uses a type of Markov chain to identify a discrete set of moments in which an abort-and-retry attempt should be enacted. The work is limited in a number of ways: it only works at a discrete set of moments; if there is an anomaly, the task must be restarted from the beginning; and it only discerns between success and failure and not between other modes of nominal or anomalous behavior. 
In \cite{2013IROS-Nakamura-ErrorRecTaskStrat}, Nakamura \et present a theoretical error recovery system that works across types of manipulation classes. The paper organizes manipulation tasks hierarchically with primitives in the bottom, and compound tasks higher up. The work suggests recovery solutions through forward or backward correction steps. Backward correction steps revert execution to the beginning of one of the hierarchical layers. Forward corrections execute minimal adjustments that help finalize a manipulation step. The theoretical work is useful, particularly when there is a well defined graph of behaviors for a task. However, no experimental work was offered in this work. 
In \cite{2013ICAR-Chang-BotTaskRecovery_PetriNets_LfD}, Chang \et devised an error recovery system based on Petri Nets learned from demonstration. Error conditions are defined based on object location: if objects are not located in expected states, an error is triggered. An interesting aspect of the work is that recovery is learned from a human demonstrator. The downside is that the system needs to maintain a growing list of expected object locations. The work does not consider errors that arise from other causes. 

The listed works assume that anomalies are caused by internal representation errors, \ie: sensing, modeling, and planning. While these error types are certainly relevant, robots now also face a threat from external disturbances such as unintended collisions with human partners, objects, or the environment. Such anomalies may even lead to a further presence of anomalies.  

In \cite{2013IROS-DiLello-BayesianContFaultDetection}, DiLello \et used a non-parametric Bayesian Hidden Markov Model in an alignment task to identify specific failure modes when extraneous objects where placed in the workspace preventing the robot to achieve a proper alignment. His work showed the identification of failure modes using wrench signals. The work however did not attempt recovery measures. In our work, we also use non-parametric Bayesian techniques with multi-modal signals, but in our case, we develop robust recovery techniques across tasks and conditions. The work of Park \et in \cite{2016ICRA-Park-MultiModalMonitoringAnomalyDet_RobotManip} studied the effects of multi-modal sensory signatures in a hidden Markov model (HMM) for anomaly identification. Their work identified anomalies in pushing tasks (doors and switches) and feeding tasks. The anomaly threshold was updated according to the progress of the task, but the work did not test any kind of recovery. As in this work, we too use Bayesian priors, but we make use of a nonparametric form that allows us to learn the complexity of each mode according to the data, allowing us to generate more expressive identification models which directly affects our task recognition and recovery rates \cite{2017humanoids-rojas-shdp-var-hmm}. 
The work of Salazar \et in \cite{2017Science-Salazar-Correcting_Robot_Mistakes_in_Real_Time_using_EEG_Signals} introduced anomaly recovery by using human mind signals in real-time to alert the robot if it had made a mistake. The work used EGG Error-Related Potential (ErrP) signals as well as secondary interactive error-related potential signals that further alerted the robot if the human caught a second mistake. The approach is compelling as the human is able to influence the robot's behavior but also the robot influences the human behavior. The work did not study how to help the robot learn from experience. One of our goals is to grow a library of motion/identification models that the robot accumulates over time to learn new behaviors. 

Finally, the work in \cite{2015RSS-Kappler-DateDrivenOnlineDecisionMakingManipu} devices a supervised machine-learning framework for online decision making in manipulation tasks. The system closes a loop between a high-level decision making system with a low-level loop. The high-level system makes use of two classifiers to identify nominal behaviors and failure. The system learns new skills online, including recovery skills and is able to save them as Dynamic Motion Primitives (DMPs) in the low-level layer. This work advanced the state-of-the-art significantly by integrating robot introspection, failure characterization, decision making, and anomaly recovery. However, the work did not provide quantitative results for recovery and only showed one recovery for one task at one moment in the task. We are interested in further studying the robustness of recovery behaviors. That is, can a system recover multiple times from disturbances? Can it do so at different points in a task or across multiple tasks? Our goal is effective recovery from disturbances at any location in the task any number of times.
\section{Problem Formulation} \label{sec:prob_formulation}
\begin{figure}     
	\centering		
		\includegraphics[scale=0.9]{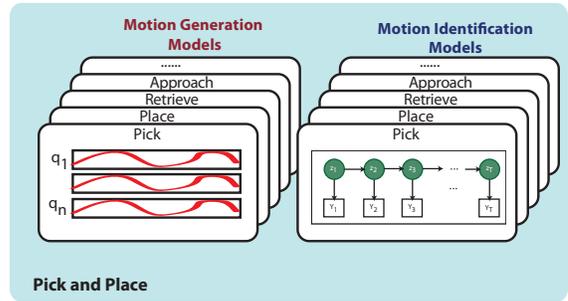}
		\caption{A motion library example composed of motion generation and motion identification models. The number of models in the library can grow over time. Motions can be designed or learned according to the preferred motion generation scheme. Identification models record sensory-motor patterns for training. Fig. \ref{fig:framework} shows an illustration of the task graph, where nodes call specific motion and identification models.}
        \label{fig:library}        
\end{figure}

Motion's inherent structure is composed of a sequence of primitive or compound skills ${S_m}$ similar to that of language grammar \cite{2017iros-rojas-onlinewrenchintrospection,2015RSS-Kappler-DateDrivenOnlineDecisionMakingManipu,2006CAS-LinHeger-TowardsAutomaticSkillEvaluation:DetectionSegmentationRobotASMotion,2006IEEETBiomedEng-Rosen-GeneralizedApproachModelingMIS_DiscreteMarkov}. Just as grammar has rules and order, motion is also organized by rules and order that yield discernible patterns in the sensory-motor action space. 
Based on this premise, we use a directed graphical model \GC \gap composed of  tasks \BC, which are interconnected by edges \EC\gap such that $\GC:\{ \BC,\EC \}, \BC=\{1,..,,B_E\}, \EC_{s,t}=\{(s,t):s,t\in \BC\}$.
Behaviors in turn consist of nodes \NC\gap and edges \EC\gap, such that: $\BC=\{ \NC,\EC \}$. Nodes can be understood as phases of a manipulation task. In our work, we prefer to name them milestones $\NC_i=(1,...,N_I)$, as they indicate particularly important achievements in a task. Nodes are composed of nodes with dual roles to define \textit{motion generation} and \textit{motion identification}. Any task graph is bootstrapped by a simple linear structure that grows as more skills and identification models are learned over time. Motion generation modules can be encoded by any given motion generation algorithm (smooth joint-interpolation, motion planning, or point attractor systems \cite{2012AutBot-Rojas-AutHetBotAsmbly,2013NC-Ijspeertdynamical-CMP_LrnAttracMdls_MtrBeh} or Probabilistic encoding \cite{2016Arxiv-Meier-ProbabilisticDMPs}). In this work, motion interpolation is used to encode skills necessary for various tasks like pick-and-place and open-and-close-a-drawer. Motion identification uses Bayesian non-parametric Markov (switching) models \cite{2017humanoids-rojas-shdp-var-hmm} to learn nominal models and consequently generate corresponding failure identification (Sec. \ref{sec:robot_introspection}). Fig. \ref{fig:library} illustrates a library with motion generation and motion identification modules, both of which get called simultaneously by nodes in the graph. 

During node execution, robot introspection uses multi-modal robot sensory inputs to build nominal skills models and then define failure identification models (Sec. \ref{sec:robot_introspection}). We limit ourselves to failure identification and do not attempt failure classification (which remains an open research question as the variability and number of possible failures can be prohibitively large in unstructured scenarios). The recovery system is triggered when an anomaly is detected, at which point, the strategy reverts the task to a node determined based on task dependencies (Sec. \ref{sec:recovery}). The dependency is set by humans and indicates a stable node in the graph where a skill can be re-issued such that the reverting can be safe and stable. Finally, the introspection system is tightly connected to the lower-level system, which executes the arbitration: running the current skill and its successor, or the recovery skill and the appropriate skill after reverting. The framework is illustrated in Fig. \ref{fig:framework}.
\section{Motion Skills}\label{sec:motor_skills}
Motions are encoded using Dynamical Motion Primitives (DMPs) \cite{2009ICRA-Pastor-LfD_DMP}. The DMP framework uses a set of nonlinear differential equations whose point attractor system is defined by a nonlinear forcing function, and which in turn depends on a canonical system for temporal scaling. The derivation of DMPs can be found in \cite{2009ICRA-Pastor-LfD_DMP} and is not included here for brevity sakes. Motion skills are trained as individual skills and stored within the motion module of a node in the task. As with Associative Skill Memories \cite{2015RSS-Kappler-DateDrivenOnlineDecisionMakingManipu}, sensory-motor experiences are used to learn introspection models \cite{Rojas_n_Peters:2005:ABBI,2015RSS-Kappler-DateDrivenOnlineDecisionMakingManipu} necessary for anomaly identification.
\section{Robot Introspection} \label{sec:robot_introspection}
The robot introspection model uses non-parametric Bayesian priors along with a Hidden Markov model (HMM) and either Gaussian or Autoregressive emission models. 
First, HMMs are a doubly stochastic and generative process used to make inference on temporal data. The underlying stochastic process contain a finite and fixed number of latent states or modes $z_t$ which generate observations $Y=\{ y_1,...,y_t \}$ through mode-specific emission distributions $b(y_t)$. These modes are not directly observable and represents sub-skills or actions in a given node of a task. Transition distributions, encoded in transition matrix $A_{ji}$, control the probability of transitioning across modes over time. The model assumes conditionally independent observations given the generating latent state. Given a set of observations, the Baum-Welch algorithm is used to infer model parameters $\Pi=(A,b)$. The fixed-modes assumption in HMMs limits the model's expressive power as it is unable to derive natural groupings. Bayesian nonparametric priors extend HMM models to learn latent complexity from data \cite{2010-Fox-BNP_LearningMarkovSwitchingProcesses,2014AOAS-Fox-JointModel_MutlipleTimeSeries_BetaProcess_MotionCapture}. We use Fox's \et sticky-Hierarchical Dirichlet Process (sHDP) prior with an auto-regressive switching system \cite{2010-Fox-BNP_LearningMarkovSwitchingProcesses} to model nominal skills as in our previous work \cite{2017humanoids-rojas-shdp-var-hmm} and derive more robust failure identification methods in manipulation tasks, specially in recovery scenarios. 

Bayesian statistics are combined with the sHDP prior to both learn model complexity $k$ from the data but also to model the transition distribution of an HMM. The sticky parameter in the prior discourages fast-mode-switching otherwise present. Consider a set of training exemplars $\XC_t=\{ x_1, ,..., x_T\}$ of observed multi-modal data $\tau$ consisting of Cartesian pose and wrench values. Then, a mode-dependent matrix of regression coefficients $\mathbf{A}^{(k)}=[A_1^{(k)} \cdots A_r^{(k)}] \in \mathbb{R}^{d\times(d*r)}]$ with an $r^{th}$ autoregressive order and $d$ dimensions is used along with a measurement noise $\Sigma$, with a symmetric positive-definite covariance matrix. The generative model for the sHDP-AR-HMM is summarized as: 
\begin{eqnarray}
	G_0=\sum_{k-1}^\infty \beta_k \delta_{\theta_k} & \beta | \gamma \sim GEM(\gamma). 					& \nonumber \\
													& \theta_k | G_0 \sim G_0. 							& \\
    G_j = \sum_{k-1}^\infty \pi_{jk} \delta_{\theta_k} 	& \pi_j | \alpha,\beta \sim DP(\alpha,\beta). 	& \nonumber
\end{eqnarray}
The probability measure $G_j$, which models the transition distribution of the modes $\pi_j$ determines the weights (probabilities) of transitioning between modes $\delta_{\theta_k}$. To avoid fixing the mode complexity, $G_j$ uses a prior $G_0$ that is unbounded and can grow with the complexity of the data. While $G_j$ uses the same set of modes as $G_0$, $G_j$ introduces variations over those points. $G_0$ provides support for a possibly infinite space, but due to the Dirichlet's process properties (\ie the Chinese Restaurant Process), a finite set of modes is selected. In fact, we can understand the hierarchical specification as $G_j=DP(\alpha,G_0)$. 

Different observation models can be included into the HMM. A Gaussian distribution with different covariance models (full, diagonal, and spherical) are considered. For Gaussian models, mode specific means and standard deviations are used $\theta_{z_t}=\mathcal{N}(\mu,\sigma^2)$. 
Additionally, the sHDP-HMM can be used to learn VAR processes, which can model complex phenomena. The transition distribution is defined as in the sHDP-HMM case, however, instead of independent observations, each mode now has conditionally linear dynamics, where the observations are a linear combination of the past $r$ mode-dependent observations with additive white noise. A prior on the dynamic parameters $\{ A^{(k)}, \Sigma^{(k)} \}$ is necessary. A conjugate matrix-normal inverse Wishart (MNIW) was used to this end. The generative process for the resulting HDP-AR-HMM is then found in Eqtn \ref{eqtn:gen_proces_shdp_var_hmm}.
\begin{equation}
    \mbox{Observation Dynamics: }\\ 
    	y_t=\sum_{i=1}^r A_i^{(z_t)} y_{t-i} + e_t(z_t). \nonumber
        \label{eqtn:gen_proces_shdp_var_hmm}
\end{equation}
\begin{equation}
	e_t \sim \mathcal{N}(0,\Sigma).
\end{equation}
\begin{equation}
	\mbox{Mode Dynamics: }\\
    	z_t^{(i)} \sim \pi_{z_{t-1}^{(i)}}^{(i)}. \nonumber
\end{equation}
By using the model over a set of multi-modal exemplar data $\XC_t$, the sHDP-AR-HMM can discover and model shared behaviors in the data across exemplars. Scalable incremental or ``memoized" variational coordinate ascent, with birth and merge moves \cite{2014AOAS-Fox-JointModel_MutlipleTimeSeries_BetaProcess_MotionCapture} is used to learn the posterior distribution of the sHDP-HMM family of algorithms along with mean values for the model parameters $\theta$ of a given skill, hence $\theta_{S_m}=\{\Pi,\textbf{A}\}_{S_m}$.
\subsection{Anomaly Identification} \label{subsec:identification}
The robot introspection system simultaneously detects nominal skills and anomaly events. Models are trained for individual skills to capture dynamics from multi-modal observations through vector $\tau_m$. $\tau_m$ consists of end-effector pose and wrench values. Scalable incremental or ``memoized" variational coordinate ascent, with birth and merge moves \cite{2014AOAS-Fox-JointModel_MutlipleTimeSeries_BetaProcess_MotionCapture} is used to learn the posterior distribution of the sHDP-HMM along with mean values for the model parameters $\Pi$ of a given skill $s$. Hence $\Pi_s=\{\pi,\textbf{A}\}$: the transition matrix and regressor coefficients.
\subsubsection{Nominal Classification} \label{subsubsec:nominal}
Given $S$ trained models for $M$ robot skills, scoring is used to compute the \textit{expected cumulative likelihood} of a sequence of observations $\mathbb{E} \left[ log \mbox{ } P(Y | \Pi_s) \right]$ for each trained model $s \in S$. 
Given a test trial $r$, the cumulative log-likelihood is computed for test trial observations conditioned on all available trained skill model parameters $log \mbox{ } P(y_{r_1:r_t} | \Pi)_s^S$ at a rate of 200Hz (see Fig. \ref{fig:likelihoods} for an illustration). The process is repeated when a new skill $m$ is started. Given the position in the graph $s_c$, we can index the correct log-likelihood $\mathbb{I}(\Pi_s=s_c)$ and see if its probability density of the test trial given the correct model is greater than the rest: 
\begin{eqnarray}
	log \mbox{ } P(y_{r_1:r_t} | \Pi_{correct}) > log \mbox{ } P(y_{r_1:r_t} \given \Pi_s)v\nonumber \\
	\quad \forall s(s \in S \land s \neq s_c). \nonumber
    \label{eqtn:state_classification_condition}
\end{eqnarray}
If so, the identification is deemed correct, and the time required to achieve the correct classification recorded. At the end of the cross-validation period, a classification accuracy matrix is derived as well as the mean time threshold value (these results were reported in \cite{2017humanoids-rojas-shdp-var-hmm}, in this paper we limit ourselves to report on the recovery robustness of the system). 
\subsubsection{Anomalous Identification} \label{subsubsec:anomalous}
\begin{figure}[t]     
	\centering		
		\includegraphics[scale=1.]{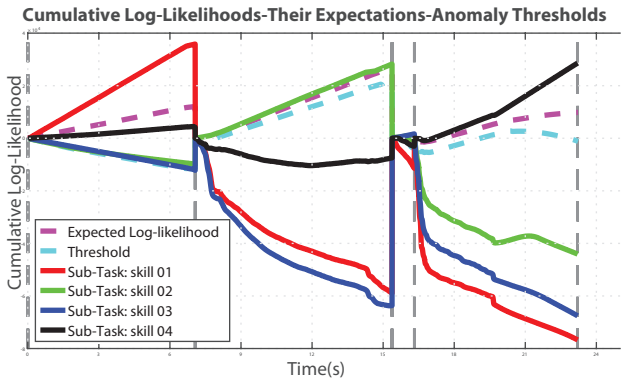}
		\caption{Illustration of cumulative log-likelihood (L-) curves for four skills $s$ in a task. As such, 4 probabilistic models $\Pi_s$ are derived. Given an indexed position in graph $N_i$, one can generate four L-curves. The curve for the correct model, has a value increasing likelihood, while the other graphs have value decreasing likelihoods. One can also see how at the termination of a skill, all data is reset and restarted. No anomalies are shown in this plot.}
        \label{fig:likelihoods}        
\end{figure}

Anomaly detection assumes that the cumulative log-likelihood $L$ of a set of nominal skill exemplars $\XC_{\SC}$ share similar cumulative log-likelihood patterns. If so, the expected cumulative log-likelihood derived in training can be used to implement an anomaly threshold $F1$. Initially, we consider a likelihood curve generated from training data for a given skill $s$. Then, for each time step in an indexed skill $s_c$, the anomaly threshold is set to $F1_{s_c}=\mu(L)-k*\sigma(L)$, where $k$ is a real-valued constant that is multiplied by the standard deviation to change the threshold. Here, we are only interested in the lower (negative) bound. Then, an anomaly is flagged if the cumulative likelihood crosses the threshold at any time: $\mbox{if } log \mbox{ } P(y_{r_1:r_t}  \given \Pi_{correct}) < F1_{s_c} \mbox{: anomaly, else nominal}$. In Fig. \ref{fig:likelihoods}, note the 4 probabilistic models. Given an indexed position in the graph, an anomaly threshold corresponding to that skill's expected cumulative log likelihood. The figure also illustrates how at the end of a skill, all data is reset and restarted. 

\begin{figure}[b]     
	\centering		
		\includegraphics[scale=0.7]{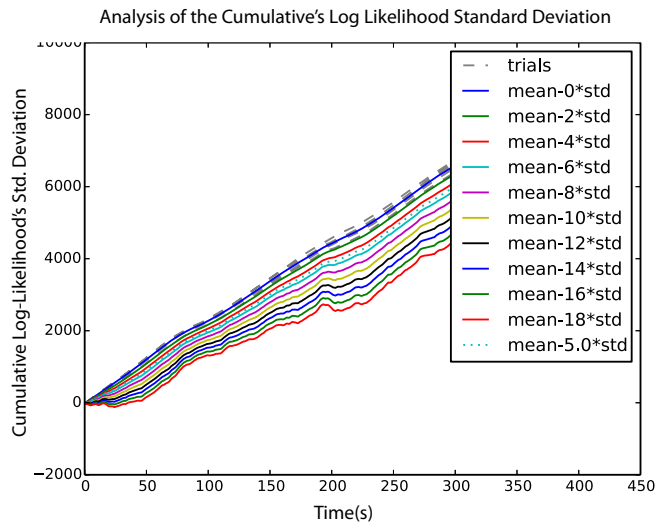}
		\caption{This plot shows how the standard deviation of the cumulative log-likelihoods computed during training grows over time. The standard deviation grows as observations show greater variance due to the accumulation of error.}
        \label{fig:growing_std_dev_likelihood}        
\end{figure}

\begin{figure}     
	\centering		
		\includegraphics[scale=0.9]{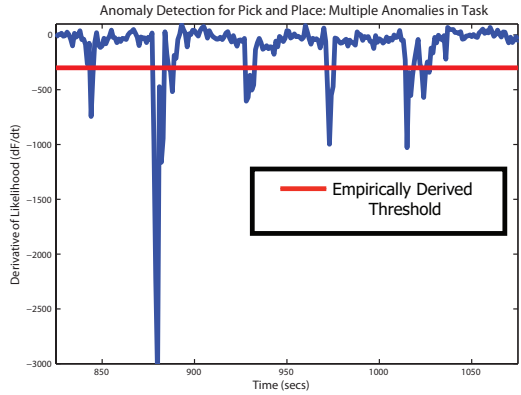}
		\caption{Example of how the anomaly identification metric crosses the anomaly threshold as anomalies are triggered by external perturbations during a skill execution (Sec. \ref{subsubsec:anomalous}).}
        \label{fig:derivative_threshold}        
\end{figure}

Upon our initial exploration of recovery schemes we noticed that after resetting the cumulative log-likelihood observations in the HMM model, false-positive anomalies were triggered at the beginning of the skill. Further examination revealed that the standard deviation of cumulative log-likelihood graphs during training began with small variances but grew over time as shown in Fig. \ref{fig:growing_std_dev_likelihood}). Given that variances are small at the beginning of the task, small variations in observations can trigger failure flags. 
A second threshold definition was designed to overcome this situation and used in our work instead of F1. As the difference between $L$ and $F1$ is minimal at first the new anomaly threshold $F2$ (for an indexed skill) is focused on computing the derivative of the difference: $F2_{s_c}=\frac{d\mid L-F1_{s_c} \mid}{dt}$. Fig. \ref{fig:derivative_threshold}, illustrates the derivative signal crossing the empirical anomaly threshold as anomalies are triggered by external disturbances.
\section{Anomaly Recovery} \label{sec:recovery}
The Anomaly recovery policy is generic such that unique policies can be easily and flexibly used across tasks. Designing proper recovery policies is challenging. A robot system as part of an unstructured environments must understand not only its state (what he is doing), but also the state of the world and that of the objects of interest, \ie how should I respond when the state of the objects of interest and the world change? Decision making may change across different anomaly types, different tasks, and even robots of different morphology. 

Our recovery scheme is designed to revert the task execution to a stable skill (or node) in the task and then re-attempt the execution of a behavior with updated goal parameters. The policy does not simply replay a action. The replaying of a skill is not a passive action, instead the skill encodes the manner of performing an action, and it requires current goals to execute properly. The key is to determine how many skills to revert? 
We use a state-dependency criteria that determines if a current skill depends on a previous one for safe execution. Safe execution is defined as \textit{giving the robot a pose and skill that can overcome the external disturbance currently experienced}. Dependencies in nodes are currently annotated manually. Dependencies are also recursive. Consider a pick action fails due to a perturbation. Re-attempting the skill directly would likely fail as the end-effector pose or target object location might have been modified. The system identifies that the pick node has a dependency pointing to the pre-pick node and so reverts to that target node. The system then examines if the pre-pick node has a dependency, if it does it would revert again; otherwise, the system resumes task execution.

In conclusion, the system flows as follows: If the system state is nominal, the current skill executes until termination and transitions to the successor. If an anomaly is detected, a state-dependency is extracted and reverted is enacted. The new skill issues low-level control commands to try to overcome the current disturbances. Note that, currently, during the recovery stage--in which the manipulator returns to the goal pose of a previous skill--the robot introspection is shut-down preventing us to introspect at this stage. This is left as future work. 
\section{Experiments and Results} \label{sec:experiments}
Three manipulation experiments are designed to test the robustness of the online decision making system for anomaly recovery. We use accidental human collisions as disturbances for a pick-and-place and open-and-close-drawer tasks, and tool collisions for a pick-and-place task. For each of the three experiments, we test four separate conditions: (i) no anomalies, (ii) anomalies without recovery, (iii) one anomaly caused at each executed skill and (iv) multiple anomalies caused at each skill. The pick-and-place task consists of 5 basic nodes (not counting the home node and the recovery node): pre-pick, pick, pre-pick (returns to an offset position), pre-place, and place. The open-and-close-drawer task consists of 5 nodes: pre-grip, grip, pull-to-open, push-to-close, and go-back-to-start. The direction and intensity of the human collisions was random but all executed under the sense that these are accidental contacts as a human user reaches into the workspace of the robot without noticing the robot's motion. We note that while the tasks are simple, the main focus of the work is placed on the robustness of the recovery systems given different external disturbance scenarios.
A dual-armed humanoid robot -Baxter- was used. All code was run in ROS Indigo and Linux Ubuntu 14.04 on a mobile workstation with an Intel Xeon processor, 16GB RAM, and 8 cores. 

For motion generation, two techniques were used for the different tasks. For the pick-and-place task, we used Baxter's internal joint trajectory action server which uses cubic splines for interpolation. The goal target is identified online through image-processing routines. For the open-and-close-a-drawer task, we trained dynamic movement primitives using kinesthetic teaching \cite{2013NC-Ijspeertdynamical-CMP_LrnAttracMdls_MtrBeh}.

In terms of robot introspection, the sHDP-HMM code with ``memoization'' variational coordinate ascent, with birth and merge moves was implemented using BNPY \cite{bnpy} and wrapped with ROS. Training used 10-trial batches. Observations used 13 dimensional vectors composed of 7 DoF pose values (position and quaternion) and 6 DoF wrench values. A baseline HMM algorithm was implemented through HMMLearn \cite{hmmlearn} and wrapped with ROS. 
The anomaly threshold for each skill was computed through leave-one-out cross-validation.

Fig. \ref{fig:exp_setup_pp_anomaly}, shows a representative image of the Baxter robot attempting a pick operation. A human collaborator accidentally collides with robot before a pick action. Note that collisions were strong enough to to move the current pose significantly from the intended path and sometimes collide with other parts of the environment. The robot introspection system identifies an anomaly and triggers a recovery behavior. The lower left part of the image shows the anomaly F2-metric flagging the anomaly. The system then begins recovery as seen in the directed graph on the right (implemented in ROS-SMACH). Video and auxiliary data for the three experiments under the four conditions are available in \cite{2017Humanoids-Rojas-supplementalURL}.
\begin{figure*}[thb]     
	\centering		
		\subfigure{\label{fig:exp_setup_pp_anomaly}\includegraphics[scale=1]{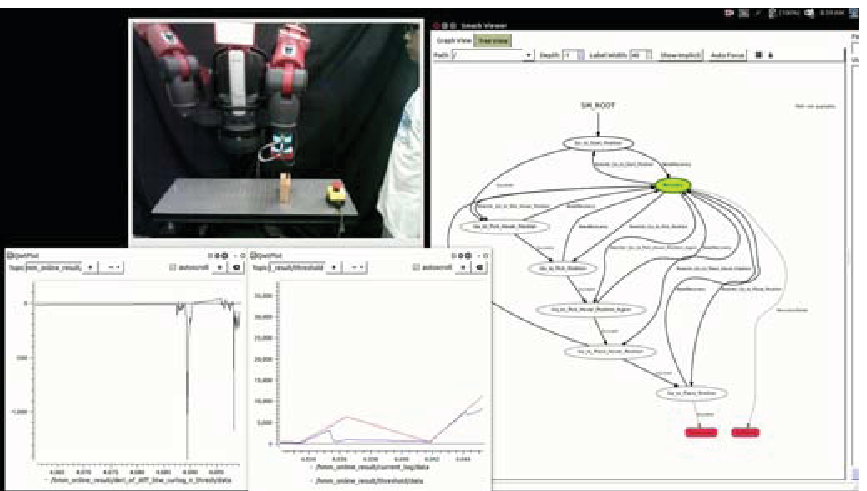}}
        \subfigure{\label{fig:exp_setup_od_anomaly}\includegraphics[scale=1]{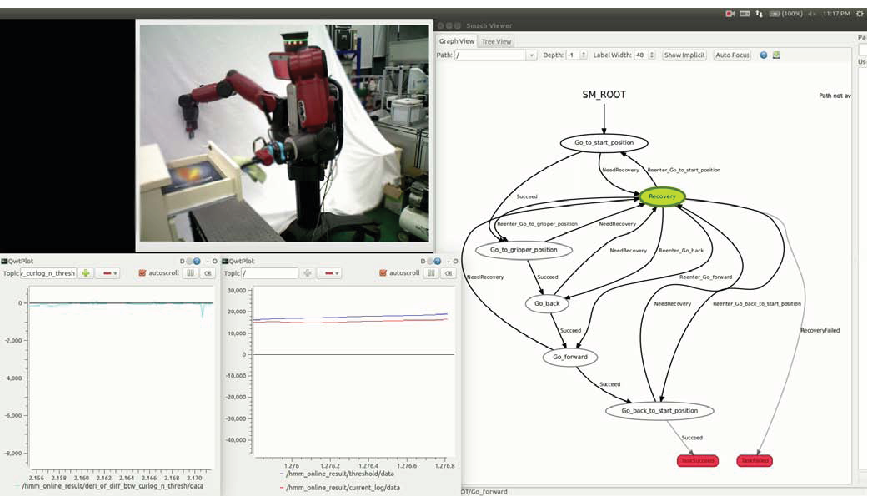}}
		\caption{Two examples (pick-and-place and open-and-close-drawer) in which a human collaborator accidentally collides the robot. The introspection system identifies an anomaly (see bottom left plots) and triggers a recovery behavior (see the fluorescent node in the graph on the right).}
        \label{fig:exp_setup_pp_anomaly}        
\end{figure*}
For results reporting, two markers are provided: (i) A success recovery rate for the recovery policy and (ii) an F-score, precision, and recall numbers for assessing anomaly identification. In particular, these markers are used under  experimental conditions (iii) and (iv) conditions described at the beginning of Sec. \ref{sec:experiments}. The first two conditions are use as a baseline and compare with the generic recovery behavior success rates. \\\\
\newcounter{Experiment} 
\stepcounter{Experiment} 
\theExperiment) {Human Collisions}\\
\textbf{Condition 3: One anomaly per node}: 27 test trials were used for the pikc-and-place task and 24 test trials were used for the open-and-close drawer task. Given that both of the tasks have 5 nodes, a total of 5 anomalies were induced in the task, 1 per node. Table \ref{tbl:results}, shows the recovery success rate of the task (represented by the F-score) and the robustness of the identification system through the precision-and-recall quantities for both micro and macro settings. The pick-and-place recovery had an average success rate of 85\% with a maximum of 88\%. The precision was 100\% (for both macro/micro) indicating strong resistance to false positives. The recall was $\sim$82\% (for macro/micro) indicating the existence of some false-negatives. This might indicate that there might have been some collisions that were of lower magnitudes than the ones we might have trained with and were not detected by the system. For the drawer task, the recovery success rate was 91.67\%. The precision was $~$95\% (macro/micro) and recall was $\sim$84.5\%. As with the human collision, weaker contact signals in tests compared to training might be the reason for the presence of false-negatives in our system. \\

\textbf{Condition 4: Multiple anomalies per node}: Under this scenario the pick-and-place task ran 42 test trials and the drawer task ran 30 test trials. Five anomalies were induced repeatedly one-after-the other for each node. The pick-and-place recovered 85\% of the time with a precision of $\sim$95\% and a recall of $\sim$84.5\% (micro/macro). The drawer task recovered $93.3\%$ of the time with a precision of $\sim$100\%, and a recall of $\sim$93\% (micro/macro).\\\\
\stepcounter{Experiment}
\theExperiment) Tool Collisions\\
For the tool collision experiment, we only tested recovery in the pick-and-place task. The results for this scenario were similar to that of human collision. The recovery success rate was $\sim$88.89\%, the precision 100\%, and the recall 88.89\%. 
\begin{table*}[b]
  \centering
  \caption{Results for accidental human and tool collisions. Recovery strategy success rates are presented by the F-score, and robustness of the robot introspection system for anomaly is shown in the Recall and Precision settings. We also compare the performance of the more expressive sHDP-HMM model with an HMM that serves as a baseline. Generally, the sHDP-HMM model allowed for better introspection and thus better recovery rates and better recall.}
  \label{tbl:results}
\begin{tabular}{lccccccc|ccccc}
               & \multicolumn{1}{l}{} & \multicolumn{5}{c}{HMM}                                         & \multicolumn{1}{l|}{} & \multicolumn{5}{c}{sHDP-HMM}                                                 \\
\midrule               
              					& \multicolumn{1}{l}{} 			&         & \multicolumn{2}{c}{Micro} & \multicolumn{2}{c}{Macro} 			& \multicolumn{1}{l|}{} & \multicolumn{1}{l}{} 	& \multicolumn{2}{c}{Micro} & \multicolumn{2}{c}{Macro} \\
               					&                      			& F       & Recall     & Precision    & Recall     & Precision    			&                      	& F                    	& Recall     				& Precision    & Recall     & Precision    	\\
Human Collision 				&                      			&         &            &              &            &              			&                      	&                      	&            				&              &            &              	\\
\midrule
\textbf{Pick \& Place}  		&                      			&         &            &              &            &              			&                      	&                      	&            				&              &            &              	\\
(1 An. / skill)					&					       		& \red{88.00\%} & 88.00\%    & 95.65\%      & 88.00\%    & 96.67\%      	&                      	& 82.00\%           	& 81.48\%        			& 100\%        & 82.00\%    & 100\%        	\\
(Mult. An. / skill) 			&         						& 84.06\% & 82.93\%    & 97.14\%      & 83.50\%    & 97.14\% 				&                      	& \red{85.00\%}        	& 84.09\%     				& 94.87\%      & 85.06\%    & 95.28\%     	\\
\textbf{Open Drawer}    		&                      			&         &            &              &            &              			&   					&                      	&            				&              &            &              	\\
(1 An. / skill)					&        						& 80.00\% & 80.00\%    & 100\%        & 80.00\%    & 100\% 					&                     	& \red{91.67\%}         & 90.00\%    				& 81.82\%      & 90.00\%    & 85.33\%      	\\
(Mult. An. / skill) 			&         						& 72.34\% & 71.79\%    & 100\%        & 72.34\%    & 100\% 					&                     	& \red{93.33\%}         & 93.33\%    				& 100.00\%     & 93.33\%    & 100.00\%    	\\\\
Tool Collision 					&                      			&         &            &              &            &              			&   					&                      	&            				&              &           	&              	\\
\midrule
\textbf{Pick \& Place}  		&                      			& 71.43\% & 71.43\%    & 100\%      & 71.43\%     & 100\% 					&             			& \red{88.89\%}        	& 100\%        				& 88.89\%     	& 100\%    	& 88.89\%  
\end{tabular}
\end{table*}
\section{Discussion} \label{sec:discussion}
This work showed the ability of a system to recover from unmodeled and accidental external disturbances that can't be anticipated. Such disturbances will be more common in shared human-robot workspaces. Our work demonstrated that our recovery strategy in connection with our previous introspection framework recovered 88\% of the time from accidental human and tool collisions under single-anomaly and multiple-anomaly scenarios per node. The results indicate the system can recover at any part of the task, even when it is abused and multiple anomalies occur consecutively. From the videos in \cite{2017Humanoids-Rojas-supplementalURL}, we can see that even when in cases where the robot is in constant duress, the robot recovers consistently. Such robustness will play a role in enabling robots have increasing levels of long-term autonomy.

Not many works have explored the subject of recovery with real robots in unstructured environments under the presence of significant and varied external perturbations. In \cite{2015RSS-Kappler-DateDrivenOnlineDecisionMakingManipu,2012Humanoids-Pastor-TowardsASMs}, there are examples of recovery from external disturbances, but no attempt is done to quantitatively assess the extent to which the recovery system might work. 

We by using a more expressive model to do robot introspection, our recovery ability also increased. We will continue to explore improved models that can better capture spatio-temporal relationships of high-dimensional multi-modal data. As well as looking for representations that scale over time in order to acquire a useful and practical library of skill identification and motion generation.

Yet there is much work to be done. Manual annotations for state-dependency are an important weakness. Crucially we wish to move towards modeling human understanding for decision making in the midst of robot-object-environment interactions. Scalability is an important factor in this domain as the system must scale to ever growing number of learned tasks. Learning how anomalies and recovery decisions are made and re-used across similar scenarios will be important. The manual approach will not scale. Adaptability and not only reverting is also important. Incremental learning is also key. The current work is limited to reverting. By simply revering we don't model recovery behaviors and also do not learn how to adapt. We also cannot handle new scenarios. We must develop action-confidence metrics that let us learn new scenarios on demand. 
\section{Conclusion} \label{sec:conclusion}
This work presented a robust and generic online manipulation recovery system that handles external disturbances. Robust anomaly identification is required to assist the system in flagging anomalies in uncertain and unstructured environments (typical in human-machine interaction). Our system leveraged non-parametric Bayesian HMMs to train a robust anomaly identification metric, which when flagged, triggers the recovery system. The recovery system uses a revertive policy based on a state-dependency criteria that selects a previous skill from a task-graph along with updated goals to overcome the external disturbances. 
\section{Acknowledgements} \label{sec:Acknowledgements}
This work is supported by ``Major Project of the Guangdong Province Department for Science and Technology (2014B090919002), (2016B0911006) and by the National Science Foundation of China (61750110521).''
\bibliographystyle{IEEEtran}
\bibliography{IEEEabrv,Xbib}
\end{document}